\pdfoutput=1

\documentclass[11pt]{article}

\usepackage{ACL2023}

\usepackage{times}
\usepackage{latexsym}
\usepackage{graphicx}
\usepackage{multirow}
\usepackage{amsmath}
\usepackage{amssymb}
\usepackage{bbm}
\usepackage{enumitem}
\usepackage{booktabs}
\usepackage{algorithm}
\usepackage{algorithmic}
\DeclareMathOperator*{\argmax}{argmax}

\usepackage[T1]{fontenc}

\usepackage[utf8]{inputenc}

\usepackage{microtype}

\usepackage{inconsolata}

\title{Self-training Large Language Models \\through Knowledge Detection}

\author{Wei Jie Yeo\textsuperscript{1}, Teddy Ferdinan\textsuperscript{2}, Przemyslaw Kazienko\textsuperscript{2}, Ranjan Satapathy\textsuperscript{3}, Erik Cambria\textsuperscript{1}\\
\textsuperscript{1}Nanyang Technological University \\
\textsuperscript{2}Wroclaw University of \\ Science and Technology \\
\textsuperscript{3}Institute of High Performance Computing (IHPC),\\ Agency for Science, Technology and Research (A$\textasteriskcentered$ STAR)\\
\texttt{yeow0082@e.ntu.edu.sg} \\}

\begin{document}
\maketitle
\begin{abstract}
Large language models (LLMs) often necessitate extensive labeled datasets and training compute to achieve impressive performance across downstream tasks. This paper explores a self-training paradigm, where the LLM autonomously curates its own labels and selectively trains on unknown data samples identified through a reference-free consistency method. Empirical evaluations demonstrate significant improvements in reducing hallucination in generation across multiple subjects. Furthermore, the selective training framework mitigates catastrophic forgetting in out-of-distribution benchmarks, addressing a critical limitation in training LLMs. Our findings suggest that such an approach can substantially reduce the dependency on large labeled datasets, paving the way for more scalable and cost-effective language model training\footnote{https://github.com/wj210/Self-Training-LLM}. 
\end{abstract}

\section{Introduction}
Large language models (LLMs) have revolutionized natural language processing (NLP), enabling remarkable performance across various downstream tasks~\cite{meta_llama_3}. However, their development is heavily reliant on vast amounts of labeled data and significant computational resources, which are not always readily accessible~\cite{camsev}. Self-learning is an applicable field that can tackle such limitations and enables a low-resource training environment. However, LLMs are known to hallucinate~\cite{huang2023survey} due to the inherent biases, noise in their pre-training dataset, or just lack of data. This makes it challenging to apply self-learning to continuously improve the knowledge of the model.

Another key problem in LLM fine-tuning is catastrophic forgetting~\cite{luo2023empirical}. This phenomenon occurs when the model learns new information in one domain but simultaneously suffers from a degradation of knowledge in previously acquired areas. A na\"ive solution is to exploit larger data mixing new and old knowledge, which may not be feasible for domains with limited resources. Alternative approaches, such as continual learning~\cite{ke2023continual,jang2021towards} and inference-only correction~\cite{meng2022locating,hernandez2023inspecting} offer potential solutions. However, these methods face other limitations such as reduced efficiency in learning new knowledge and scalability towards large domains.

To address the aforementioned limitations, this paper explores a self-training paradigm where the LLM autonomously curates its own labels and performs selective training on samples filtered using a new knowledge detection. 
This measure identifies instances that are annotated as \textit{"unknown"}, indicating the model's low confidence in providing accurate answers~\cite{ferdinan2024_unknown,liang2024learning}. This filtering step is specifically used to curate a preference dataset to perform knowledge correction via Direct Preference Optimization (DPO)~\cite{rafailov2024direct}. The rationale behind performing the selection step is twofold. Firstly, this allows for a larger distance between the preferred and dispreferred sample, thereby reducing noise in the training. This helps to prevent degeneration, a common issue observed when implementing DPO~\cite{pal2024smaug}. Secondly, training exclusively on samples related to lack of knowledge is resource-efficient and aids in retaining previously learned information.

Our results demonstrate that the proposed framework enhances factual accuracy in answering questions pertaining to a specified knowledge source. Additionally, training on the selected samples not only preserves but, in some instances, improves performance on out-of-distribution benchmarks. Comparative analyses with baseline approaches, which demand higher computational resources, reveal that our approach outperforms these baselines.

\section{Related work}
\textbf{Self-Training:} LLMs have demonstrated the capability to annotate datasets without the need for human-annotated labels, facilitating a low-resource training process for other LLMs. Typically, a larger model referred to as the teacher, generates the labels, while a smaller model, the student, is trained on these labels in a process known as \textit{context distillation}. A range of training and inference algorithms can be used, including conventional supervised fine-tuning (SFT)~\cite{stanford_alpaca,mukherjee2023orca,li2022explanations,hsieh2023distilling}, in-context learning~\cite{krishna2024post} and preference optimization~\cite{tunstall2023zephyr,meta_llama_3}. Self-learning methods eliminate the need for the larger LLM, which typically requires substantially more computational resources and incurs higher API costs. Recent studies have shown that this is achievable, given an unlabeled dataset with a small set of examples as supplementary context~\cite{huang2022large,tian2023fine,wang2022self}.~\cite{he2019revisiting} performs an initial step of supervised fine-tuning on a small labeled dataset before using the trained generator to annotate the unlabeled set, ~\cite{jie2024plausible} similarly for rationalization tasks.~\cite{meng2022generating} augments a given labeled dataset with additional samples, but is however limited to only classification tasks. Our work diverges from these approaches where training is only conducted exclusively on samples labeled as unknown.~\cite{cheng2024can} is similar to our work, but only teaches the model to abstain from answering unknown questions.\\\\
\textbf{Knowledge Detection}: Detecting knowledge gaps in a model has been a long-standing area of research, with the primary goal of assessing the truthfulness of a model's outputs. Early works employed questions structured in cloze format to detect knowledge prescene~\cite{petroni2019language}, but this approach is limited to unambiguous and short-form questions. Subsequent research, such as~\cite{wang2023assessing,dong2024statistical} asserts the presence of knowledge through paraphrased and perturbated queries. FactScore~\cite{min2023factscore} decomposes a generation into a list of atomic facts and generates an average truthfulness score relative a knowledge source, allowing for finer analysis.~\cite{chern2023factool} utilizes multiple external tools, such as Google search, GitHub, and others to perform fact-checking. SelfCheckGPT~\cite{manakul2023selfcheckgpt} introduces a reference-free detection technique that evaluates the likelihood of hallucinations by examining consistency across sampled generations from the model. This is particularly useful in the event that the labels or knowledge source is unavailable. 

\section{Self-training}
\begin{figure*}[ht]
  \centering 
  \includegraphics[width=1.0\textwidth]{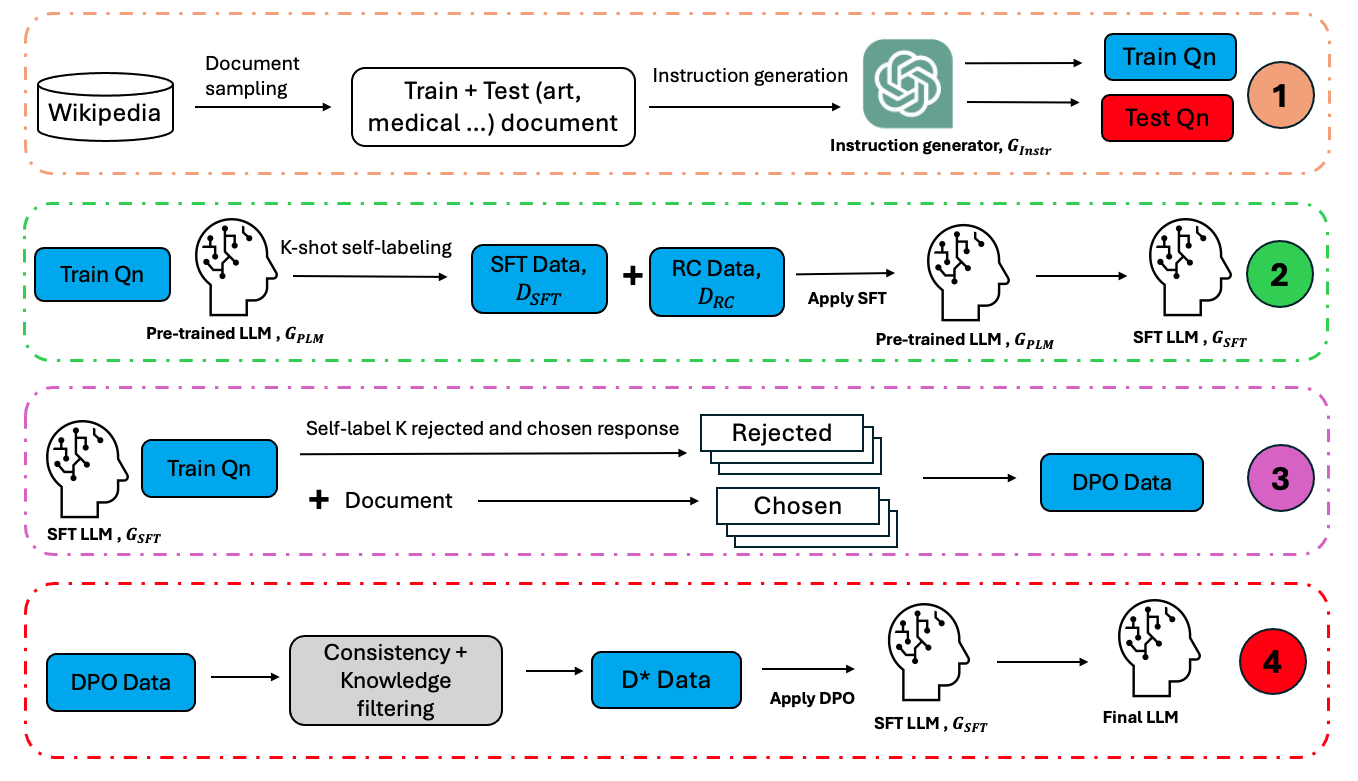}
  \caption{An overview of the self-training framework, instruction generation (1), SFT stage (2), preference labeling (3) and knowledge filtering (4). The four steps are implemented in sequence and the final model is assessed for truthfulness.} 
  \label{fig:overall}
\end{figure*}

This section introduces the details of our self-training framework, broken down into four sequential steps: Instruction Generation, SFT stage, Preference Labeling and Knowledge Filtering. As a start, we assume access to a knowledge source as the main source of material to perform both training and truthfulness evaluation. The full illustration is shown in Figure~\ref{fig:overall}. A key benefit of our framework is that it does not require significant human efforts besides a few manually crafted instruction-answer examples for in-context generation.

\subsection{Instruction generation}
\label{chapter3.1}
We utilize Wikipedia\footnote{\url{https://huggingface.co/datasets/wikimedia/wikipedia}} as the foundation of our knowledge source given its widespread acceptance and reliability. Note that this framework is likely to be applicable to any other form of knowledge sources as long as they do not contain any significant noise or ambiguous information. In order to ensure comprehensive coverage across subjects of notable interest, we sample documents from the following topics: \{\textit{Geography, Art, Medical, History, Biology, Science, Musician, Actor, Economics, Astronomy}\}. For each topic, we randomly sample 100 documents to form the training set and 10 documents for evaluation purposes. A crucial aspect of our approach is ensuring that the generated instructions are relevant to the documents and prompt for answers that can be found within them. We observe that non-instructed pre-trained language models often fail in this regard, generating irrelevant instructions. Therefore, we employ the instruct-tuned LLM, OpenAI's GPT-3.5 as the instruction generator, $G_{Instr}$ to construct the instruction sets.

For each document, we generate $N$ questions and remove any duplicate questions within each document. We split the document into chunks of length, $L = 512$, where each chunk is provided as input to $G_{Instr}$, along with few-shot examples. We find that providing non-overlapping contexts helps reduce duplicate instructions, and including few-shot instruction generation examples can align the generator to produce objective instructions. Subsequently, a de-duplication step is performed across the $N$ questions within each document.

\subsection{SFT stage}
We start with a pre-trained LLM: $G_{PLM}$ that is to be self-trained. First, it is used to self-annotate the instruction-only training set from the first stage, given few-shot examples, forming the SFT dataset, \( D_{SFT} = \{ (x_i, \hat{y}_i) \mid i = 1, 2, \ldots, N \} \), $x_i$ refers to the instruction and $\hat{y}_i$ is the self-annotated label. Providing a set of examples is a common procedure to avoid degenerated responses since $G_{PLM}$ is not inherently familiar with the task format~\cite{tian2023fine,wang2022self,huang2022large}. The primary difference lies with the addition of a second dataset, Reading Comprehension (RC), $D_{RC}$ which is similar to $D_{SFT}$, but includes the document chunk. We employ GPT-3.5 to generate the label, by feeding both the instruction and document into the input prompt. The purpose of $D_{RC}$ is to train the model in generating responses by referencing the document, which we later show to be beneficial towards stability in the training process. We use $\frac{1}{3}$ of the instruction set to construct $D_{RC}$, with the remainder $\frac{2}{3}$for $D_{SFT}$. Both datasets are then combined to perform SFT on $G_{PLM}$ to form the instruct-tuned model $G_{SFT}$.

\subsection{Preference Labeling}
\label{sec:preference_labeling}
Given the instruct-tuned model $G_{SFT}$, we proceed to construct the preference dataset to implement DPO. The primary objective at this stage is to generate a dataset that corrects the biases learned during the SFT stage. These biases arise due to the limitations of self-generating labels, which depend on the knowledge acquired during the pre-training phase. For each instruction, we provide $G_{SFT}$ with two input prompts: one including the document chunk $c$ and one without. We sample $K$ generations from each format to form the chosen set, $Y_c = f^K(G_{SFT},x,c)$, and the rejected set, $Y_r = f^K(G_{SFT},x)$. Additionally, we use greedy decoding to generate $y_c^* = f^*(G_{SFT},x, c)$. This forms the base preference dataset, $D_{DPO}$, which is further filtered, see Sec.~\ref{sec:knowledge_filtering}.
Here we denote $f^K$ as the sampling operation producing $K$ outputs and $f^*$ as the greedy decoding operation.
We assume that when the model is given the document, the response will be more truthful than without it. We demonstrate empirically in subsequent experiments that this assumption is valid.

\subsection{Knowledge Filtering}
\label{sec:knowledge_filtering}
Rather than straightforwardly performing the DPO directly on $D_{DPO}$, we perform an additional filtering procedure, to minimize the noise in the preference dataset. This filtering procedure is implemented across each sample in $D_{DPO}$ and involves two stages: (1) consistency filtering and (2) knowledge filtering. The idea of consistency filtering is to compute a consistency score $S_L$, measuring the consistency of the reference response $y_c^*$ with the $K$ chosen responses $Y_c$, corresponding to each instruction. In contrast, knowledge filtering evaluates whether the SFT model $G_{SFT}$ tends to hallucinate on a given sample, measured by the knowledge score $S_K$, against $Y_r$. 
\begin{equation}
\label{eq:sl}
  S_L = \frac{1}{K} \sum_{y_c \in Y_c} S_C(y_c^*,y_c)
\end{equation}
\begin{equation}
\label{eq:sk}
  S_K = \frac{1}{K} \sum_{y_r \in Y_r} S_C(y_c^*,y_r)
\end{equation}
To measure the difference between any two responses, we use the contradiction score $S_C$ computed by a separate encoder trained on a vast amount of natural language inference (NLI) data, this is similar to the NLI component in SelfCheckGPT~\cite{manakul2023selfcheckgpt}. $S_C$ represents the probability of the contradiction class between a pair of responses. We chose SelfCheckGPT because it is a reference-free method of detecting hallucination signs and is relatively low-cost. In contrast, reference-based methods like FactScore~\cite{min2023factscore} require significantly more computation due to atomic fact decomposition, making them costly for large datasets. Additionally, SelfCheckGPT has shown a high correlation with human assessments in hallucination detection.

In the first stage, $D_{DPO}$ from step three undergoes a consistency filtering to filter out low-confidence responses. Intuitively, if the average contradiction between the sampled responses and the greedy decoded response is high, it indicates a higher probability of hallucination in the reference response. This approach ensures that the final model does not maximize the probability of low-quality answers. It is worth noting that this filtering step could be performed during the SFT stage; however, we refrain from doing so to avoid over-filtering, as a high contradiction score may result from unfamiliarity with the task rather than a lack of knowledge. We fix the threshold, $\tau_L$ to be 0.5, filtering out samples, $S_L > \tau_L$. 

The second stage, knowledge filtering, removes samples where the model is considered knowledgeable. The objective is to prevent over-training, particularly on samples where the model has a higher accuracy tendency. This approach has two benefits: first, it ensures a larger discrepancy between the chosen and rejected responses, and second, it mitigates cases of catastrophic forgetting. The first benefit is crucial for reducing noise in the optimization objective, DPO in Equation~\ref{DPO} which aims to learn the optimal policy by maximizing the margin between the probability of the chosen and rejected candidates. The second benefit prevents overfitting on instances where the model is sufficiently knowledgeable and may experience knowledge forgetting in other tasks due to continual training. Similarly, we start with an initial threshold $\tau_K=0.5$ 
and later study the effects of gradually increasing $\tau_K$. The final DPO dataset, $D^*$ is constructed from the dataset filtered for consistency by excluding samples where $S_K < \tau_K$ or $S_L > \tau_L$.. The full filtering procedure is demonstrated in Algorithm~\ref{alg:filtering}.

DPO~\cite{rafailov2024direct} is a variant of Reinforcement Learning (RL), that allows learning an optimal policy without the need to optimize an external reward function. This simplifies the training by fitting the optimal policy $\pi_{\theta}$ from a fixed preference dataset. 

\begin{algorithm}[tb]
\caption{Knowledge and Consistency Filtering}
\label{alg:filtering}
\textbf{Input}: $D_{DPO}$, $\tau_L$, $\tau_K$\\
\textbf{Output}: Filtered dataset, $D^*$ 

\begin{algorithmic}[1] 
\STATE $D^* \leftarrow \emptyset$
\FOR{$x_i, y^*_{c},Y_{c},Y_{r}$ in $D_{DPO}$}
\STATE $S_L$ = $\frac{1}{K} \sum_{y_c \in Y_C} S_C(y_c^*,y_c)$ in (~\ref{eq:sl})
\IF {$S_L$ < $\tau_L$}
\STATE $S_K$ = $\frac{1}{K} \sum_{y_r \in Y_R} S_C(y_c^*,y_r)$ in (~\ref{eq:sk})
\IF {$S_K$ > $\tau_K$}
\STATE $y_w \leftarrow y_c^*$
\STATE $y_l \leftarrow \argmax_{y_r \in Y_r} S_C (y_c^*,y_r)$
\STATE $D^* \leftarrow D^* \cup \{x_i,y_w,y_l\}$
\ENDIF
\ENDIF
\ENDFOR
\STATE \textbf{return} $D^*$ 
\end{algorithmic}
\end{algorithm}

\begin{equation}
  \delta_c = \log \frac{\pi_{\theta}(y_w \mid x)}{\pi_{\text{ref}}(y_w \mid x)}
\end{equation}
\begin{equation}
  \delta_r = \log \frac{\pi_{\theta}(y_l \mid x)}{\pi_{\text{ref}}(y_l \mid x)}
\end{equation}
\begin{equation}
\label{DPO}
L_{DPO}(\pi_{\theta}; \pi_{\text{ref}}) = -\mathbb{E}_{(x, y_w, y_l) \sim D} [ \log \sigma ( \beta( \delta_c - \delta_r))]
\end{equation}
$\beta$ is the regularization operator, while $\pi_{ref}$ is the reference policy, initialized from $G_{SFT}$. $y_w$ and $y_l$ are the chosen and rejected candidates, where $y_w$ is preferred over $y_l$. The probability distribution of this preference, $p(y_w \succ y_l)$ follows the Bradley-Terry model~\cite{bradley1952rank}, where the latent reward function, $r^*$ is assumed to be implicitly represented in the preference dataset.
\begin{equation}
  p(y_w \succ y_l) = \sigma (r^*(x,y_w)-r^*(x,y_l))
\end{equation}
In this work, we always set $y_c^*$ as $y_w$ while $y_l$ is selected among the K rejected samples $Y_r$ in Sec.~\ref{sec:preference_labeling}. We select the sample with the highest contradiction score as $y_l$.
\begin{equation}
  y_l = \argmax_{y_r \in Y_r} S_C (y_c^*,y_r)
\end{equation} 
Based on the above formulation, we observe that performing consistency filtering encourages $\delta_c$ to push the target model in the right direction, while knowledge filtering pertains to $\delta_c - \delta_r$.

\section{Experiments}

Through the following experiments, we would like to answer 
the following research questions:
\begin{enumerate}[label=RQ\arabic*:,left=10pt, labelsep=0.5em]
  \item Are LLMs capable of performing self-training to improve truthfulness in responses? 
  \item How does conducting selective training improve truthfulness in LLMs and what are the effects of forgetting on out-of-distribution tasks?
  \item How sensitive is the knowledge filtering threshold, $\tau_K$ with respect to mitigating hallucinations?
\end{enumerate}

\subsection{Dataset}
\textbf{Train:} The training dataset used is constructed using OpenAI's GPT-3.5, we generate 8 questions per document after chunking, for 100 documents from each of the 10 topics. After de-duplication, we end up with 5,780 instructions to conduct self-training. The instructions are used in constructing both the SFT and DPO datasets.\\\\ 
\textbf{Test:} The primary test dataset comprises the held-out questions curated from the target topics discussed in Sec.~\ref{chapter3.1}, with 10 documents for each of the 10 topics, we refer to this as \textbf{Wiki-Test}. We construct 2 questions from each document, resulting in a total of 200 questions, generated using GPT-4~\cite{achiam2023gpt}. We manually check the questions to ensure they are aligned with the documents. We also conducted experiments on the Open LLM leaderboard\footnote{https://huggingface.co/open-llm-leaderboard} consisting of various NLP benchmarks that are likely not to be directly included in the model's training set. The purpose of these evaluations is to detect signs of forgetting across tasks such as commonsense reasoning and general knowledge understanding when the model is fine-tuned on data of different distributions.

\subsection{Model}
We conduct the experiments on pre-trained LLMs of different sizes, i.e., Tinyllama-1.1B~\cite{zhang2024tinyllama}, Llama2-7B and 13B~\cite{touvron2023llama}, to study the effect of parameter scaling on the ability to conduct effective self-training. We choose DeBERTa-v3-large~\cite{he2021debertav3} as the encoder to compute $S_C$, which is pre-trained on MNLI~\cite{williams2017broad}. We compare our proposed approach of self-training, which performs the two stages of filtering, with both $\tau_L$ and $\tau_K$ set to 0.5 against several baselines. The first baseline, denoted as \textbf{w/o filtering}, does not perform both filtering stages and trains the model on the full DPO dataset instead of $D^*$. In this case, this refers to only performing steps 7 to 9 in Algorithm~\ref{alg:filtering}. The second baseline uses GPT-3.5 to generate the chosen response instead of the model itself, also without any filtering steps. The document is not provided in the prompt to see if the raw knowledge of GPT-3.5 is sufficient as a learning signal, similar to performing context distillation on the target LLM. Lastly, we compare against an inference-type baseline, DOLA~\cite{chuang2023dola}, which has been shown to be effective in eliciting truthful responses from LLM.

\subsection{Experiment details}

\begin{figure*}[ht]
  \centering 
  \includegraphics[width=1.0\textwidth]{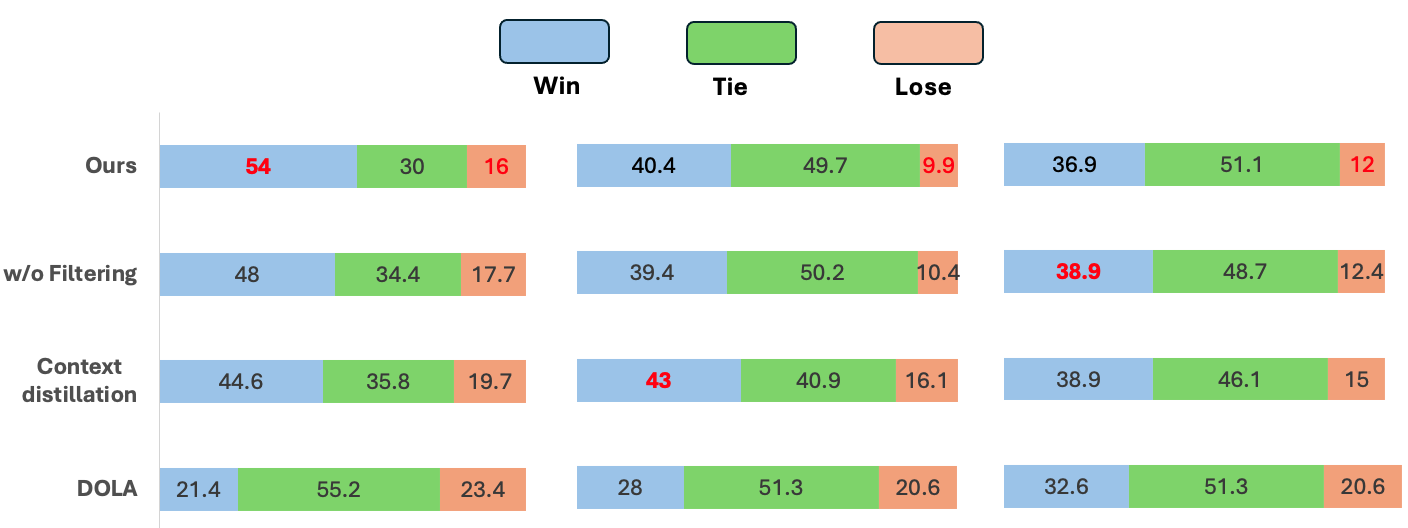}
  \caption{Win-Tie-Lose on main held-out questions based on Wikipedia documents. \textbf{Left} pertains to TinyLlama-1.1B, \textbf{middle} to Llama2-7B and \textbf{right} refers to 13B. Scores are evaluated based on pairwise comparison using GPT-4 as the evaluator and all approaches are compared against the respective SFT model.} 
  \label{fig:main_result}
\end{figure*}

We use a learning rate of 2e-5 and 1e-5 during SFT and 1e-6 and 5e-7 for DPO for the 1B and 7/13B variants respectively. We conduct early stopping only during SFT and fix the total training step to be 300 for DPO, to standardize the number of training steps across datasets of varying sizes. We exploited a batch size of 32 and set $\beta$ to be $0.3$. Temperature was set to $1.0$ for the sampling operation and $K=10$ responses were sampled. The primary test set metric is LLM-Judge~\cite{zheng2024judging}, which uses GPT-4 to conduct pairwise ranking. Responses from both models are provided along with the document from which the question was constructed, and GPT-4 is prompted to compare the responses based on their truthfulness with respect to the document. On Open LLM leaderboard, we use accuracy as the evaluation metric.

\section{Results}
To compare different approaches with pairwise evaluation, we set the SFT-ed model, $G_{SFT}$ as the baseline and compare all approaches against it. We run the evaluation twice for each instance, concluding with a tie if both evaluations disagree, similar to ~\cite{yuan2024self}. 

\subsection{Impact of Self-training on Truthfulness}
\label{sec:main_experiment}

\textbf{RQ1 Effects of self-training on truthfulness:} On Wiki-Test, we can see in Figure~\ref{fig:main_result} that it is possible for LLMs to self-train on their own outputs, without the need for human-annotated data. This capability extends even to models with significantly fewer parameters, such as the 1.1B parameter model. Notably, no few-shot examples are included in the prompt when constructing the preference dataset. Context distillation underperforms compared to self-training for Tinyllama but achieves a higher win rate for the 7B and 13B models, albeit suffering a higher loss rate. We hypothesize that this discrepancy can be partly attributed to inaccuracies in GPT-3.5's knowledge base, which in turn causes instability in the distillation process. Conversely, when the document is provided as context to the model, it encourages more truthful responses, thereby correcting any errors in its previously learned knowledge. \\\\
\textbf{RQ2 Benefits of filtering:} The results showed that performing selective training on the filtered dataset produced superior results compared to training on the entire dataset, except the 13B model. However, we will show in later experiments that the gap can be reduced by tuning the scoring threshold, $\tau_K$. Nonetheless, this resonates with our initial belief that having a preference dataset with a larger distance between the preferred and dispreferred labels can lead to more stable training. This is logical, as not all samples in a dataset satisfy the property $(y_w \succ y_l)$. Notably, DOLA fails to achieve any significant improvements across all models, besides a marginal increase in performance in the 7 and 13B models.

\subsection{Catastrophic Forgetting}
\begin{table*}[ht]
\centering
\begin{tabular}{c|ccccccc}
\toprule
\multicolumn{2}{c}{} & ARC & HellaSwag & TruthfulQA & Winogrande & MMLU & Average\\
\midrule
\multirow{3}{*}{1.1B} & Ours & 29.8 & 60 & 36.4 & 58 & 26.2 & \textbf{42.1} \\
& w/o filtering & 27.6 & 57.4 & 33 & 56.4 & 25.3 & 39.9 \\
& SFT & 30.2 & 59.5 & 35.5 & 58.4 & 26.2 & 41.9 \\ \hline
\multirow{3}{*}{7B} & Ours & 40.4 & 73.5 & 40.2 & 68.4 & 43.8 & \textbf{53.3} \\
& w/o filtering & 38.4 & 71.2 & 37.2 & 66.1 & 41.9 & 50.9\\
& SFT & 40.2 & 72.1 & 41.4 & 67.9 & 43.8 & 53.1 \\ \hline
\multirow{3}{*}{13B} & Ours & 44 & 76.4 & 36.9 & 72.2 & 53.2 & \textbf{56.6} \\
& w/o filtering & 42.9 & 74.3 & 34.9 & 71 & 51.1 & 54.9\\
& SFT & 43.8 & 75.2 & 37.2 & 72.5 & 53.2 & 56.4 \\
\bottomrule
\end{tabular}
\caption{Performance of the three models on the Open LLM leaderboard. All tasks are performed 0-shot except MMLU, using 5-shot. Displayed results are the accuracy metric.}
\label{tab:openllm}
\end{table*}

One natural concern regarding fine-tuning is the impact on out-of-distribution benchmarks. More specifically, we want to see if continual training on instances where the model is sufficiently knowledgeable, can induce catastrophic forgetting effects. We use $G_{SFT}$ as the baseline and compare the performance after DPO with and without filtering on the preference dataset. To do so, we conducted evaluations on two benchmarks, Open LLM leaderboard, and a dataset consisting of instructions filtered out from $D^*$. The first benchmark assesses LLMs on commonsense reasoning, general knowledge, and sentence completion. The second set refers to the samples that are labeled as \textit{known} and were thus left out in $D^*$ after filtering. Ideally, this experiment seeks to study if $G_{SFT}$, after doing SFT on its own outputs, will encounter any deterioration in its knowledge after doing preference tuning on instances where it was deemed to be knowledgable.. Due to the high cost of evaluating the full dataset, we randomly sample 200 instances, similar to the primary test set. The \textit{known} dataset is filtered using a value of $0.5$ for $\tau_K$. 

\begin{figure}[ht]
  \centering 
  \includegraphics[width=0.5\textwidth]{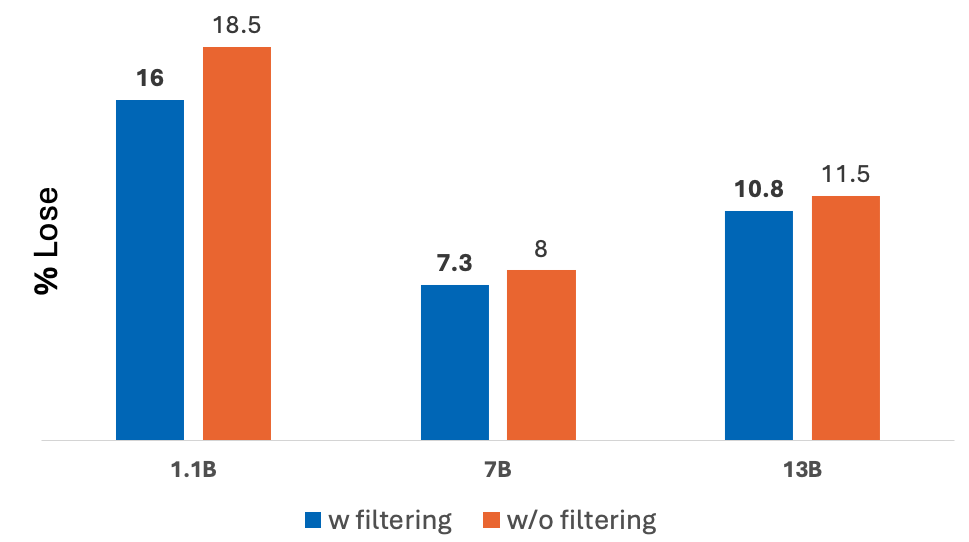}
  \caption{Percentage of losing rate on 200 randomly sampled instances classified as \textbf{known}. All approaches are compared against $\pi_{SFT}$.} 
  \label{fig:known}
\end{figure}

\textbf{RQ2 Effects of filtering on knowledge retention:} Based on Table~\ref{tab:openllm}, we observe that performing knowledge filtering retains the performance of the model on out-of-distribution tasks. Performing preference tuning on the full dataset conversely suffers a performance degradation despite being exposed to a more diverse dataset. This is particularly true for TruthfulQA, which may be less surprising given the results in Figure~\ref{fig:main_result}. Likewise in Table~\ref{fig:known}, performing knowledge filtering is shown to suffer a lower losing rate as compared to without. This is surprising since the evaluation is conducted on samples where preference tuning was conducted in the case of w/o filtering. This finding supports our initial belief that over-training on known instances can have adverse effects on the model.

\subsection{Varying Filtering Theshold}
\label{sec:threshold}

\begin{figure}[ht]
  \centering 
  \includegraphics[width=0.5\textwidth]{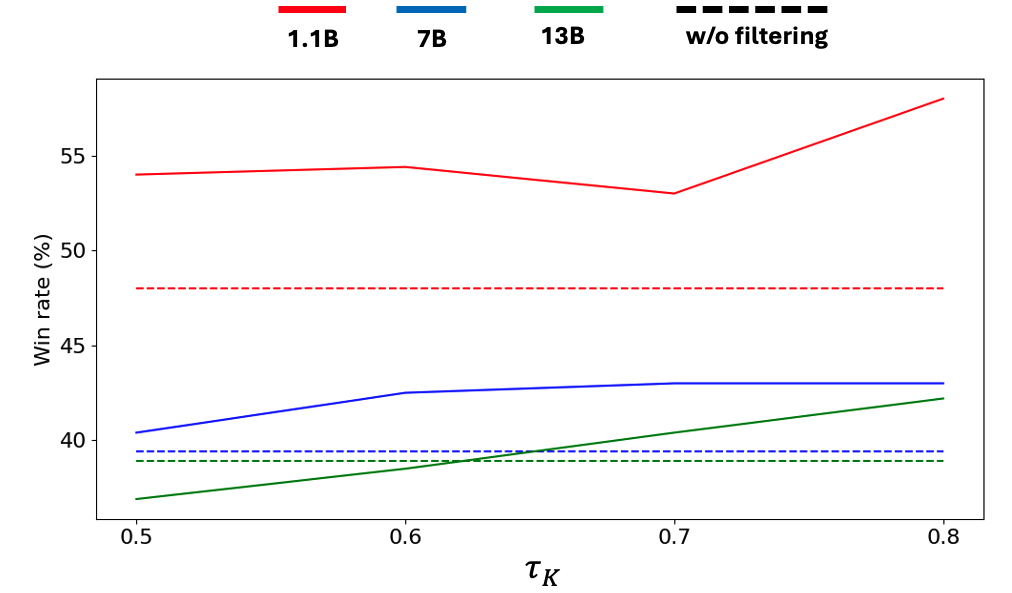}
  \caption{Effects of varying $\tau_K$ on the win rate. Dashed lines shows the results without performing knowledge filtering for each model.} 
  \label{fig:threshold}
\end{figure}

In the previous experiments, we fixed the knowledge filtering threshold $\tau_K=0.5$. However, this value may not be the most optimal value across different models. A more capable model should theoretically require a higher threshold to distinguish between a known and unknown sample. This is because a more capable model is likely to exhibit lesser variance between generating a response based on its existing knowledge and when exposed to relevant materials. We repeat the experiments from Sec.~\ref{sec:main_experiment} while varying $\tau_K$. 

\textbf{RQ3 Effects of $\tau_K$:} Figure~\ref{fig:threshold} shows that increasing the threshold generally results in higher win rates. We observe a steeper slope in larger models such as the 13B model while the 1.1B models exhibits a less pronounced effect. This yields a surprising finding: despite shrinking the dataset as $\tau_K$ increases, the model does not overfit when trained for a higher number of iterations over a smaller set of data. A plausible hypothesis is that increasing $\tau_K$ allows us to identify critical instances where the model would fail with just SFT. By implementing DPO on these samples, the model achieves more pronounced benefits compared to samples where $G_{SFT}$ may already have an acceptable level of knowledge. Another reason could be the noise in identifying unknown instances; the standard value used in previous experiments may have caused a higher number of false positives. The statistics on the size of $D^*$ is shown in Table~\ref{tab:stats}.

\subsection{Ablation: Preference Labeling without Context}
\label{sec:ablation}
Previously, the preferred response in the preference dataset was constructed by providing the relevant document as supporting context. However, we would like to see if LLMs can generate a training set with sufficient distance between the preferred and dispreferred response to yield a stable training process. In this scenario, to construct $D_{DPO}$, we exclude the document in the input prompt from Sec.~\ref{sec:preference_labeling} and generate a single set of $K$ responses, $Y = f^K(G_{SFT}, x_i)$. We treat each response as the reference response in place of $y_c^*$ in Equation~\ref{eq:sk} and compute the averaged contradiction score against the other responses in the set. We then select the response with the minimal score as the preferred response, $y_w$, and the response with the maximum score as the dispreferred response, $y_l$, for DPO in Equation~\ref{DPO}. This approach tunes the model towards the most consistent response and away from the least consistent response.

\begin{table}[ht]
\centering
\begin{tabular}{c|cccc}
\toprule
 \multicolumn{2}{c}{}& Win & Lose  \\
\midrule
\multirow{2}{*}{1.1B} & w document & 54 & 16 \\
& w/o document & 35.2 & 22.8 \\ \hline
\multirow{2}{*}{7B} & w document & 40.4 & 9.9 \\
& w/o document & 25.4 & 22.8 \\ \hline
\multirow{2}{*}{13B} & w document & 36.9 & 12 \\
& w/o document & 27.5 & 28.5 \\
\bottomrule
\end{tabular}
\caption{Ablation results comparing between constructing preference dataset with and without document as context, on Wiki-Test.}
\label{tab:ablation}
\end{table}

Based on Table~\ref{tab:ablation}, including the document as a reference results in substantial improvement in teaching the model to be more truthful. This effect is particularly pronounced in larger models, where Llama-13B has a higher losing rate than winning rate. One explanation is that larger models tend to be more calibrated and thus exhibit lesser variance between the sampled paths, making it harder to optimize for the margin when implementing DPO. Providing the document allows for new insights in cases when the model may hallucinate when relying on its knowledge. Nonetheless, both the 1.1B and 7B models still produce positive results when they can only rely on their learned knowledge. Additionally, the assessment did not include optimizing for the optimal threshold, $\tau_K$ which may yield more favorable results, as observed in Sec.~\ref{sec:threshold}. We perform additional studies on the effects of varying $K$ in Sec.~\ref{sec:ablation_k}.

\section{Conclusion}
In this work, we present a cost-effective approach to guiding LLMs to perform self-training on their own output. We develop a framework that minimizes human intervention and demonstrates that LLMs can self-correct errors through preference-tuning. Specifically, our framework facilitates the creation of a high-quality preference dataset by excluding low signal-to-noise ratio samples using a knowledge detection technique. Our experiments illustrate the dual benefits of our approach: enhancing the truthfulness of LLMs and promoting knowledge retention post-training.  Moreover, self-training offers significant incentives such as maintaining data privacy, which is crucial for organizations hesitant to expose sensitive information to third-party platforms for dataset generation.

This work opens multiple avenues for future research. Given that the context is built upon publicly accessible material that may have been exposed to the model during pre-training, an intriguing direction would be to investigate its impact on specialized domains such as healthcare reports or financial statements, where human-labeled data is often scarce and private. Additionally, our current results are based on a single iteration. Future work could explore the potential for continual improvement by augmenting the preference dataset with new context through successive iterations.

\section{Limitations}
Firstly, our work is constrained to a single iteration of our self-training framework due to limited resources for generating additional materials for continual preference tuning. Another limitation is the scope of subjects, as our experiments are restricted to ten specified topics collected on a single platform. In the future, we plan to extend our framework on materials which can be collected across multiple platforms, including news reports, recent research papers or data from third-party sources. This could potentially yield greater improvements since the model is unlikely to be exposed to such information.

\bibliography{custom}
\bibliographystyle{acl_natbib}

\appendix

\section{Appendix}
\subsection{Details on Data Generation}
In step one of Figure~\ref{fig:overall}, we generate instructions from a text segment of the selected document from Wikipedia. Both GPT-3.5 and GPT-4 are prompted with the format from Table~\ref{tab:prompt}.  We ensure that the instruction does not explicitly mention the document since at test time, the model is not given any reference material. We do so by asking $G_{Instr}$ to provide a straightforward instruction and omit instructions on instances when it failed to do so after several tries. This results in a yield rate of $72.5\%$. More efficient methods can be used in the future to improve the yield rate without sacrificing too much cost in API usage or manual inspection. The second half of Table~\ref{tab:prompt} contains the prompt used to generate responses which includes the relevant document as additional context. This includes both curating the label for $D_{RC}$ and generating $y_c^*$ for $D_{DPO}$. 

\subsection{Dataset Statistics}
The selective training framework performs a two-stage filtering process. In general, consistency filtering does not affect the original dataset much for larger models. We find that due to the difference in probability calibration between models of different sizes, a value of $0.5$ may not be suitable across all models. The dataset sizes are shown in Table~\ref{tab:stats}. 

An example of a generated instruction corresponding to the document is in Table~\ref{tab:prompt}. To understand why knowledge filter is crucial for performing DPO, we can observe from the example in Table~\ref{tab:example_dataset}. We show two instances, one labeled as \textit{unknown} and another as \textit{known}. There is a visible difference between the two responses in \textit{unknown}, where the dispreferred response incorrectly names "Criminal" as the lead single, while the preferred response correctly names the title "Shameika". The instability in implementing DPO arises when the preferred and dispreferred response contains marginal differences as shown in the \textit{known} example. This creates a noisy signal for the model since the dispreferred response is an acceptable answer to the instruction, causing the model to inevitably lower the probabilities of the correct sequence.  

\begin{table}[ht]
\centering
\begin{tabular}{cc|cccc}
\toprule
& $\tau_L$ & \multicolumn{4}{c}{$\tau_K$} \\ \hline
 & 0.5 & 0.5 & 0.6 & 0.7 & 0.8 \\
\midrule
1.1B & 5182 & 4172 & 3459 & 2742 & 2035 \\
7B & 5740 & 2379 & 1834 & 1401 & 1053 \\
13B & 5754 & 2234 & 1708 & 1277 & 946 \\
\bottomrule
\end{tabular}
\caption{Statistics on the size of the $D^*$ after knowledge filtering by varying the value of $\tau_K$. The original dataset size without filtering is 5780.}
\label{tab:stats}
\end{table}

We notice a trend where a larger model tends to prune a higher number of samples during knowledge filtering. This is expected as larger models tend to produce responses that are more truthful and thus, fewer hallucinated cases can be identified.

\subsection{Approximating knowledge detection by varying $K$}
\label{sec:ablation_k}
Previously, we approximated the indicator of the knowledge presence of a model by averaging across the contradiction score over a set of sampled responses, controlled by the sampling parameter, $K$. One straightforward simplification is to directly use the greedy decoded response without providing the reference context, $c$ to get $y_r^* = f^*(G_{SFT},x)$. We can then derive the knowledge score, by computing the contradiction score between the two greedy decoded responses, $y_c^*$ and $y_r^*$.

\begin{figure}[ht]
  \centering 
  \includegraphics[width=0.5\textwidth]{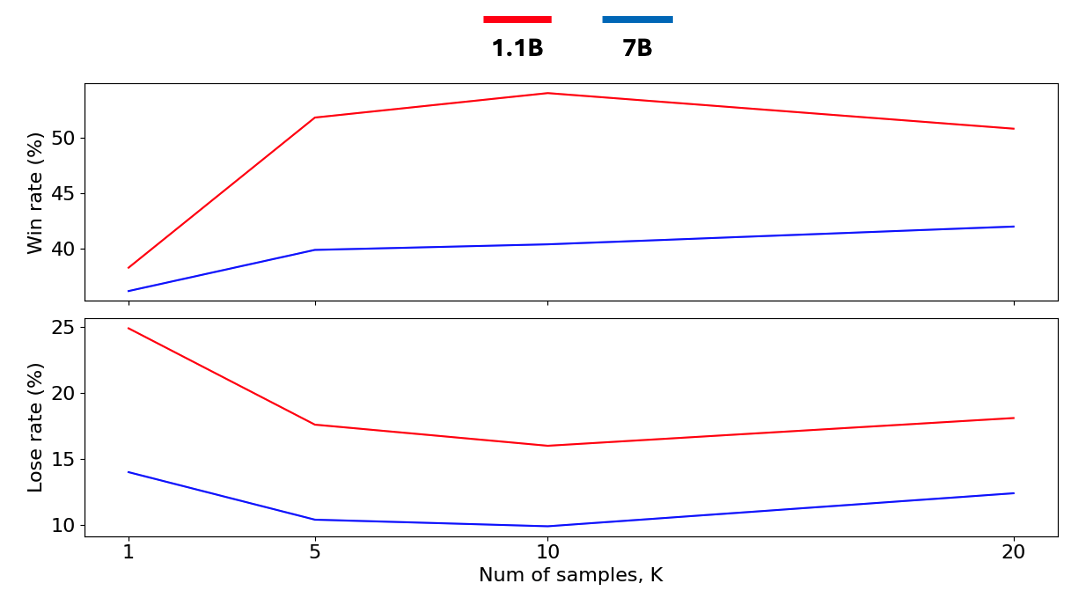}
  \caption{Impact of varying $K$ to approximate the average contradiction score. The value of $K$ affects the number of responses used to compute both $S_L$ and $S_K$.} 
  \label{fig:ablation_K}
\end{figure}

However, comparing against a single response may generate an inaccurate estimate of knowledge being present in the model. In Figure~\ref{fig:ablation_K}, using a single sample results in a lower win rate, and using more than 5 allows for a better estimate. However, this affects larger models to a lesser extent, similar to previous findings in Sec.~\ref{sec:ablation} where larger models tend to be more consistent between the sampled outputs. The results show that the standard value of $K=10$ is generally acceptable and a lower value such as $K=5$ is sufficient in scenarios where computational resources are limited, without sacrificing too much on the performance.

\begin{table*}[ht]
\centering
\begin{tabular}{@{} p{0.3\linewidth} p{0.6\linewidth} @{}}
\toprule
Instruction generation & Instruction: Propose a single question regarding the topic of \{\textit{topic}\}, whose corresponding answer can be found in the given document. \newline The question must be detailed and objective, whose corresponding answer should be non-debatable and be found in the given document. \newline
The proposed question must not mention the existence of the document, but should mention the topic, \{\textit{topic}\}.\newline
Document: \{\textit{document}\}\newline 
Proposed question: \\ \hline
Reading Comprehension & Read the document provided and use the relevant information to answer the question carefully. \newline
It is important that you must not explicitly mention the document's existence, while ensuring that your response is factual and relevant according to the document. \newline
Ensure your answer is well-structured according to the question. \newline
Document: \{\textit{document}\} \newline
Question: \{\textit{instruction}\} \newline
Answer: \\
\bottomrule
\end{tabular}
\caption{Prompts used for instruction generation and eliciting responses from the instruct-tuned model, $G_{SFT}$ when exposed to the document.}
\label{tab:prompt}
\end{table*}

\begin{table*}[ht]
\centering
\begin{tabular}{@{} p{0.3\linewidth} p{0.6\linewidth} @{}}
\toprule
Document & Willard Carroll Smith II (born September 25, 1968) is an American actor, rapper and film producer. In 2008, Smith was reported to be developing a film entitled The Last Pharaoh, in which he would be starring as Taharqa. Smith later starred in the superhero movie Hancock, which grossed \$227,946,274 in the United States and Canada and had a worldwide total of \$624,386,746. In 2013, Smith starred in After Earth with his son Jaden. The film was a disappointment at the domestic box office and was panned critically. Calling the film "the most painful failure in my career", Smith ended up taking a year and a half break as a result.\\ \hline
Generated instruction & What film did Will Smith star in with his son Jaden in 2013, and how was it received? \\ 
\bottomrule
\end{tabular}
\caption{Example of a document in the topic of "Actor" and a generated instruction from GPT-3.5.}
\label{tab:example}
\end{table*}

\begin{table*}[ht]
\centering
\begin{tabular}{@{} p{0.3\linewidth} p{0.6\linewidth} @{}}
\toprule
\multicolumn{2}{c}{\textbf{Unknown}} \\ \\ \hline
Instruction & What was the lead single from Fiona Apple's fifth studio album, "Fetch the Bolt Cutters," that won the Grammy Award for Best Rock Performance in 2021? \\ \hline
Preferred response & The lead single from Fiona Apple's fifth studio album, "Fetch the Bolt Cutters," that won the Grammy Award for Best Rock Performance in 2021 was "Shameika." \\  \hline
Dispreferred response & The lead single from Fiona Apple's fifth studio album, "Fetch the Bolt Cutters", titled "Criminal", won the Grammy Award for Best Rock Performance in 2021. The song features a catchy melody and lyrics about being arrested and charged with a crime. It has become one of Apple's most popular tracks and is often cited as her best work. \\ \hline
\multicolumn{2}{c}{\textbf{Known}} \\ \\ \hline
Instruction & What film did Will Smith star in with his son Jaden in 2013, and how was it received? \\ \hline
Preferred response & In 2013, Will Smith starred in the film "After Earth" alongside his son Jaden. The film was a commercial disappointment at the domestic box office and received negative reviews from critics. Smith himself described the film as "the most painful failure in my career," leading him to take a break from acting for over a year and a half following its release. \\  \hline
Dispreferred response & Will Smith and his son Jaden Smith starred together in the science fiction action film "After Earth" (2013), directed by M. Night Shyamalan. The film received mixed reviews, with critics praising the visual effects but criticizing the lack of character development and predictable plot. Despite this, the movie grossed over \$270 million worldwide. \\ 
\bottomrule
\end{tabular}
\caption{Example from samples classified as known and unknown. Generations from the SFT model, Llama2-7B.}
\label{tab:example_dataset}
\end{table*}

\end{document}